\documentclass{article} % For LaTeX2e
\usepackage{iclr2027_conference,times}

\usepackage{amsmath}
\usepackage{amssymb}
\usepackage{booktabs}
\usepackage{graphicx}
\usepackage{enumitem}
\usepackage{xcolor}
\usepackage{multirow}
\usepackage{array}
\usepackage{adjustbox}
\usepackage{soul}
\usepackage{url}
\usepackage{xspace}
\usepackage{hyperref}
\usepackage{fvextra}
\DefineVerbatimEnvironment{promptblock}{Verbatim}{%
  breaklines=true,
  fontsize=\footnotesize,
  frame=leftline,
  framerule=0.4pt,
  framesep=4pt,
  xleftmargin=1.25em}

\newcommand{\benchmark}{\textsc{Pitch}\xspace}

\iclrfinalcopy

\begin{document}

\title{When Should AI Chime In? \\ Benchmarking Timely Contribution of \\ Proactive Agents in Multi-Party Chat}

\author{Ruotong Wang$^{1}$, Brian S. Yu$^{1}$, Xinyi Zhou$^{2}$, Joseph Chee Chang$^{3}$, \\
\textbf{Pao Siangliulue$^{3}$, Jonathan Bragg$^{3}$, Liwei Jiang$^{1}$, Amy X Zhang$^{1}$} \\
 $^{1}$University of Washington \quad 
$^{2}$Boise State University \\ $^{3}$Allen Institute for Artificial Intelligence}

\maketitle
\lhead{Preprint}

\begin{abstract}

Proactive agents can support group collaboration by surfacing useful information without waiting to be asked, but even relevant contributions can be disruptive if poorly timed.
In this work, we isolate the timing question: given a fixed candidate contribution such as a clarification or relevant fact, when should an AI agent chime in?
We introduce \benchmark{} (\textbf{P}roactive \textbf{I}ntervention \textbf{T}iming in \textbf{CH}at)\footnote{Project website (data, annotation systems): \url{https://social.cs.washington.edu/PITCH-website/}}, to our knowledge the first benchmark with large-scale human judgments of proactive agent timing collected from participants in the same unscripted multi-party conversations they took part in.
To collect high-quality judgments, we design a novel interface in which 2-4 people chat synchronously, author candidate AI contribution messages, place their own and others' contributions into the chat history, and then deliberate as a group about the best placement for each message.
\benchmark{} contains 384 candidate contributions and 1,091 judgments, each marking an ideal turn, an acceptable window, or that a contribution should not appear, from 89 participants engaged in 35  conversations.
Evaluating 15 models across 28 configurations, we find that while models can place contributions well when given the entire conversation history, they \textbf{systematically chime in too early when asked to decide live as the conversation unfolds}: best-turn accuracy falls from 57.8\% to 19.0\%, with nearly half of placements occurring before every participant's acceptable window.
In comparison, lay humans deciding live are more accurate (23.7\% vs 19.0\%), and fewer of their placements are premature (39.7\% vs 46.4\%).
Model reasoning traces show that 93\% of intervention justifications cite relevance, while almost none consider waiting.
By providing a human-grounded benchmark, our work reveals gaps in agents' ability to integrate into group settings and points towards proactive agents that can reason explicitly about appropriate timing.

\end{abstract}

\section{Introduction}
\label{sec:pitch-intro}

\begin{figure}[h]
  \centering
  \includegraphics[width=\linewidth]{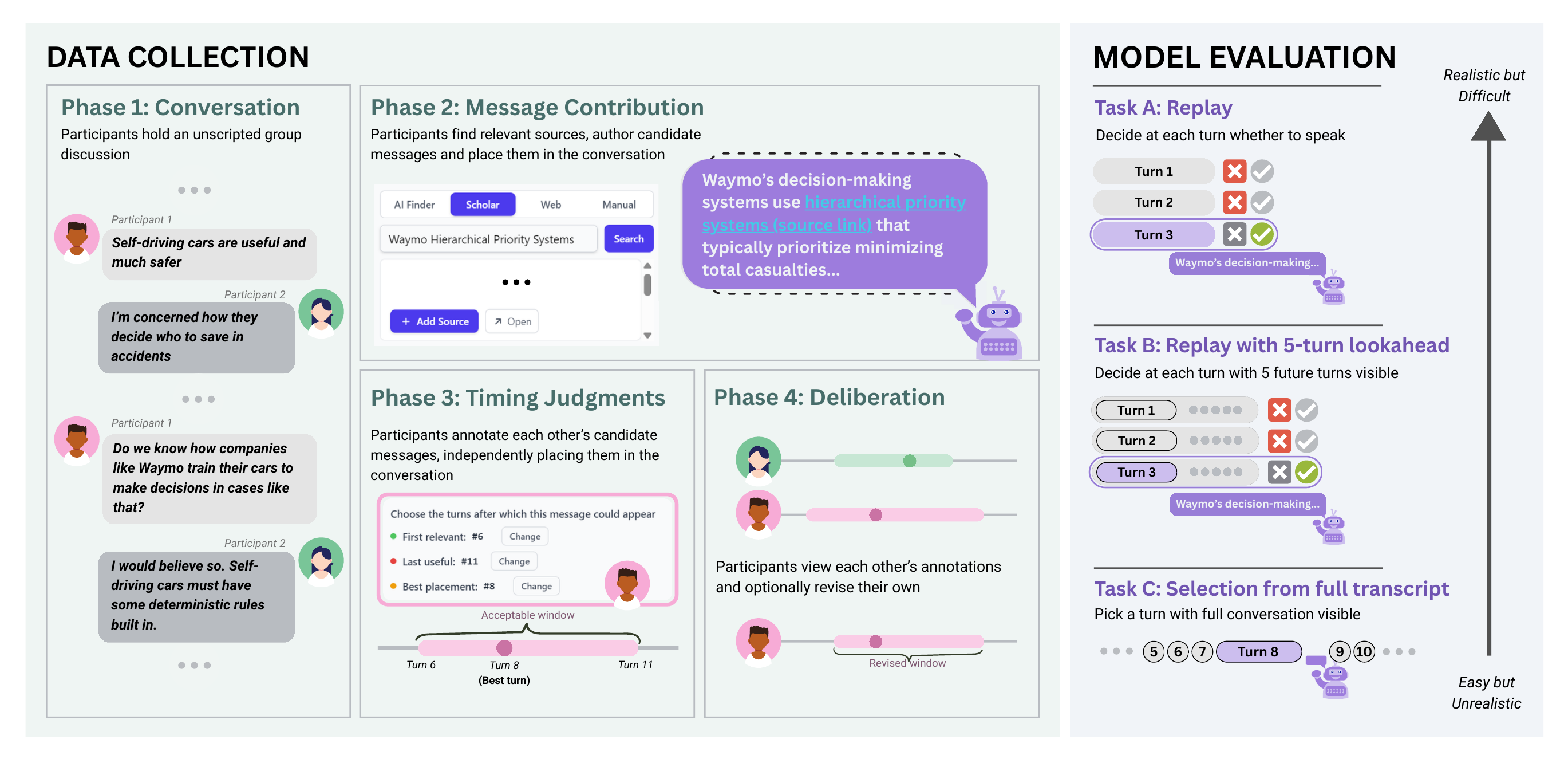}
  \vspace{-2em}
  \caption{\benchmark{} (\textbf{P}roactive \textbf{I}ntervention \textbf{T}iming in \textbf{CH}at) is the first benchmark for proactive intervention timing built from unscripted human conversations and participants’ own judgments of when proactive interjections should occur. We collected the benchmark data via a four-phrase protocol: Conversation, Message Contribution, Timing Judgments, and Deliberation. We then evaluated 15 models across 28 configurations against the benchmark in three task setups that differ in their task difficulty and realism for real-world deployment of agents in group conversations. 
  }
  \hspace{-3em}
  \label{fig:interface}
\end{figure}

Achieving proactivity in AI agents requires both having useful information to contribute and knowing when to chime in with it. Agents are increasingly capable of providing contextualized factual information such as clarifications, examples, or relevant facts that can benefit users~\cite{wang2025social}, but less attention has focused on timing.
Getting the timing right has particularly high stakes in group settings where an agent interjection can negatively disrupt the flow of conversation~\citep{yeo2026group}.
For instance, in a group chat about travel planning, an agent might have useful information about hotel pricing---but bringing it up after the group has settled on a hotel would be too late, while introducing it before they begin discussing accommodations might distract from the current topic.
Appropriate timing therefore involves weighing the benefit of contributing now against the possibility that waiting would produce a better moment~\citep{horvitz1999principles}. 

Evaluating how well agents perform on timing requires knowing when the participants in a conversation would have actually welcomed a particular contribution (e.g., a message containing useful clarifications). 
However, most prior efforts to develop proactive agents do not evaluate against this information; instead, they resort to methods that lack ecological or construct validity, including testing on synthetic conversations~\citep{chen2025llamapie}, using LLM-as-a-judge to infer timing preference~\citep{liu2025promediate}, and asking non-participants to validate by reviewing transcripts~\citep{liu2025proactive}.
Similarly, datasets developed explicitly to benchmark agent timing performance also rely on synthetic or semi-synthetic conversations with designated intervention points~\citep{patel2025discussllm,nama2026when2speak}. 
Yet work has shown that synthetic group conversations do not resemble ones between humans~\citep{kamoi2026evaluating}.
These evaluations also assume that there is only one correct placement for agent interjection and that all participants would agree on the timing. In practice, acceptable timing often spans a window and varies across participants, as our study shows.
Finally, participant preferences cannot be inferred from conversation traces alone: a transcript records what participants said, but not when they would have welcomed information that was never introduced. 

To address this gap, we introduce \benchmark{} (\textbf{P}roactive \textbf{I}ntervention \textbf{T}iming in \textbf{CH}at), which, to our knowledge, is the first benchmark for proactive intervention timing built from \textbf{unscripted human conversations and participants' own judgments of when proactive interjections should occur}. \benchmark{} is enabled by a study protocol and interactive interface designed to jointly capture a group's chat conversation, agent contribution messages, and participants' timing preferences. 
We recruit researchers and students and group each according to shared interests; they then have an unscripted synchronous chat regarding a research topic of their choosing or make shared plans for a hypothetical trip or event, respectively. Next, with the support of search and drafting tools, participants author contributions containing information that could have benefited their discussion. For each contribution, the interface  presents the message alongside the transcript and asks each group member to identify an ideal turn and an acceptable window, or indicate that the contribution should not appear at all. Participants make preferences independently before reviewing one another's placements and optionally revising their own, allowing the protocol to capture both shared preferences and disagreements that persist after deliberation.

The resulting benchmark contains 384 human-authored candidate messages from 35 conversations comprising 2{,}194 turns, with 1{,}091 timing preferences from 89 participants. For 54\% of messages, all ideal turns from participants in the group lie within two turns of one another, while 77\% have an acceptable window shared by all participants. Of the preferences revisited during peer deliberation, 85\% retain their original ideal turn. 
We therefore evaluate placements against both individual timing preferences and shared acceptable regions. These preferences provide a participant-grounded reference for the timing an online agent should anticipate, capturing both variability in individual preferences and regions of agreement across participants.

We evaluate 15 proprietary and open-weight models in 28 standard and reasoning configurations under three task setups: \textit{(A) turn-by-turn replay}, \textit{(B) replay with five-turn lookahead}, and \textit{(C) selection from the full transcript}. 
When models can inspect the completed conversation before selecting a placement (the \textit{(C) selection from the full transcript} setup), the selected placements almost always come within striking distance of participants' preferences, with 92.7\% falling within at least one participant's acceptable window, and 57.8\% directly matching at least one participant's ideal turn.
However, when models must commit without knowing what follows (the \textit{(A) turn-by-turn replay} setup), the prediction accuracy for ideal turns falls from 57.8\% to 19.0\%, and 46.4\% of predictions occur before any annotator's acceptable window opens, compared to 1.0\% after every window closes. 
We find that while non-participant lay annotators also find the online task challenging, they still perform better than models (23.7\% vs 19.0\%) and are less likely to make premature placements (39.7\% vs 46.4\%). 
When we allows models to see into the near future (the \textit{(B) replay with five-turn lookahead}), accuracy only rises to 22.4\%. 
Inspecting firing justifications made in model reasoning traces, we find that 92.8\% argue solely that the contribution is relevant to the current turn, while only 0.14\% explicitly weigh waiting for a later moment. Collectively, these results show that recognizing relevance is insufficient for well-timed proactive assistance, and that evaluating models with full conversational hindsight can substantially overstate their ability to intervene appropriately in live collaboration.

We contribute (i) \textbf{\benchmark{}, a participant-grounded benchmark of proactive intervention timings} containing human-authored contributions and large-scale timing preferences from participants in unscripted multi-party conversations (Section~\ref{sec:corpus}); (ii) \textbf{an interactive collection protocol} that elicits ideal placements, acceptable alternatives, and persistent differences in participant preferences for fixed contributions (Section~\ref{sec:study}); (iii) \textbf{a formulation and set of evaluation measures for participant-aligned timing} that preserve both individual preferences and shared acceptable regions (Sections~\ref{sec:task} and~\ref{sec:metrics}); and (iv) \textbf{empirical findings on model timing alignment}, showing a substantial gap between post hoc and online intervention decisions and models' tendency to treat relevance as sufficient reason to speak even when the timing is not appropriate (Sections~\ref{sec:groundtruth} and~\ref{sec:results}).

\section{Evaluating Proactive Agent Timing with Human Conversations and Participant-Grounded Timing Judgments}
\label{sec:study}

\benchmark{} evaluates whether models contribute messages containing helpful factual knowledge at times preferred by the \textbf{actual conversation participants}. Because these preferences are participant-specific and cannot be reliably inferred from transcripts or simulations, we designed an interactive protocol that jointly collects real conversations and participants' retrospective timing preferences.

\subsection{Four Phases of the Interactive Data Collection Protocol}
Our data collection protocol captures two key pieces of information unavailable from conversation transcripts alone: \emph{what information an AI could have proactively contributed} and \emph{when participants would have wanted to receive it}. Participants first take part in a real group discussion, then author candidate AI contribution messages containing factual information that could have proactively supported the group, such as clarifying a concept, providing evidence, or offering relevant logistical details. Using real discussions also preserves conversational dynamics that synthetic interactions may smooth away~\citep{zhou2024misleading, bian2025utopian, kamoi2026evaluating}.

Concretely, the data collection protocol consists of four phases (shown in Figure~\ref{fig:interface}):
\begin{itemize}[nosep, leftmargin=1.2em, labelsep=0.4em]
\item \textbf{Phase 1: Conversation.} Groups hold real-time, unscripted typed conversations in two settings: \emph{open research discussion}, where research literature and scientific evidence support \textit{expert} information needs, and \emph{collaborative planning} of trips or events, where prices, logistics, and options support \textit{everyday} information needs. See example topics in Appendix~\ref{sec:convo-prompts}. 
\item \textbf{Phase 2: Message contribution.} 
Participants review the conversation and author factual messages that could have helped the group, with optional search and AI support.
\item \textbf{Phase 3: Timing judgments.} 
Group members independently judge others' contributions, selecting an ideal placement, an acceptable timing window, or that the message should not appear.
\item \textbf{Phase 4: Deliberation.} 
Participants view peers' judgments and may revise their own, letting us measure whether timing preferences persist after reconsideration without requiring consensus.
\end{itemize}

This design offers two benefits. First, participant-authored contributions reflect information needs that arose in the actual discussion, making them more likely to be genuinely useful. Second, it avoids restricting the benchmark to information that current models can already identify, retrieve, synthesize, and contextualize from an ongoing group conversation.

\subsection{Task Definition: Proactive Agent Contribution Timing}
\label{sec:task}

We define a \textbf{contribution} as a message that introduces factual information intended to help the conversation, such as by clarifying a concept, providing evidence or an example, or offering relevant logistical information. In a deployed proactive-agent setting, this message would be generated by the AI agent. Given a fixed contribution, \benchmark{} evaluates \emph{when, if ever, it should be introduced}. Fixing the message isolates timing from upstream challenges such as identifying information needs, retrieving relevant information, and generating the contribution itself.

Formally, a \textbf{conversation} $C=(u_1,\ldots,u_T)$ is a sequence of turns from an unscripted multi-party human discussion, and $m$ is a participant-authored candidate contribution. The model chooses when, if ever, to insert $m$ into this fixed trace. 

For each contribution, participant $i$ provides a \textbf{timing judgment} consisting of an \emph{ideal turn} $t_i$, the best turn after which to insert the message, and an \emph{acceptable window} $[s_i,e_i]$, spanning the earliest point at which the message becomes relevant to the latest point at which it remains useful. Participants may instead select \emph{do-not-surface} if they believe the message should not appear anywhere in the conversation. Candidate contributions were reviewed for coherence and factual plausibility.

\subsection{Participant Recruitment and Study Procedures}

\noindent\textbf{Participant recruitment.}
We recruited 89 participants through social media, personal networks, and online and physical advertisements at several large research universities. Participants received a \$35 gift card for a 90-minute session. They reported topic interests in \emph{open research discussion} and \emph{collaborative planning} in advance and were grouped by shared preferences. 

\noindent\textbf{Study procedure.}
Participants first held a 15--25 minute typed conversation in groups of 2--4 using our study platform (Appendix~\ref{sec:interface-protocol}). Afterward, they reviewed the transcript and authored factual messages they believed could have helped the group, with optional search and AI drafting support. Participants were instructed to write messages that an ideal participant or agent could plausibly have contributed during the conversation. Each contribution was then evaluated by all group members, who independently selected an ideal turn and acceptable timing window or marked it \emph{do-not-surface}. This yielded 1,091 timing judgments: 384 from message authors and 707 from their 1--3 peers. Finally, participants could revisit their judgments after seeing peers' placements and optionally revise them. Of the preferences revisited during peer deliberation, 85\% retain their original ideal turn. 

\subsection{\benchmark{} Benchmark Composition}
\label{sec:corpus}

The resulting dataset comprises 35 conversations and 2{,}194 turns from 89 participants (96 group memberships), yielding 384 candidate messages and 1{,}091 timing
preferences. 
(Table~\ref{tab:conversations}). Each candidate message received a median of 3 preferences and at least 2 preferences (Table~\ref{tab:labels}).

\begin{table}[h]
  \centering
  \small
  \setlength{\tabcolsep}{2pt}
  \begin{adjustbox}{max width=\linewidth,center}
  \begin{tabular}{l  r| r | r}
    \toprule
    & \textbf{All} & \textbf{Collective Planning} & \textbf{Open Research Discussion} \\
    \midrule
    Number of conversations                                     & 35 & 14 & 21 \\
    Conversation length in turns (median, range)                     & 52 (16-227) & 79.5 (30-227) & 42 (16-92) \\
    Participant group size (median, range)                        & 3 (2-4) & 2 (2-4) & 3 (2-4) \\
    Turn length in words (median)                   & 10 & 7 & 15 \\
    \midrule
    Number of candidate messages (count)                                & 384 & 158 & 226 \\
    Candidate message length in words (median)              & 67 & 61 & 71 \\
    \bottomrule
  \end{tabular}
  \end{adjustbox}
  \vspace{-1em}
  \caption{Statistics of conversation and candidate messages of \benchmark{}.
    }
  \label{tab:conversations}
\end{table}

\begin{table}[h]
  \centering
  \small
  \setlength{\tabcolsep}{2pt}
  \begin{adjustbox}{max width=\linewidth,center}
  \begin{tabular}{l r| r| r}
    \toprule
    \textbf{Human timing labels} & \textbf{All} & \textbf{Collective Planning} & \textbf{Open Research Discussion} \\
    \midrule
    Number of participants (unique participants)                         & 96 (83) & 35 (30) & 61 (59) \\
    Timing judgment                                                & 1{,}091 & 415 & 676 \\
    Distinct participants per message (median, mean)              & 3, 2.8 & 3, 2.6 & 3, 3.0 \\
    Placements / do-not-surface            & 1{,}047 / 44 & 404 / 11 & 643 / 33 \\
    \midrule
    Gap between earliest and latest ideal turn (median / p90) & 2 / 19 & 1.5 / 14.6 & 3 / 20 \\
    Messages with a unanimous ideal turn                        & 114 (30\%) & 58 (37\%) & 56 (25\%) \\
    Individual window size (median)            & 7 & 6 & 7 \\
    \quad Anyone's best window size                                & 11 & 9 & 12 \\
    \quad unanimous window size                                    & 3 & 3.5 & 3 \\
    \bottomrule
  \end{tabular}
  \end{adjustbox}
  \vspace{-1em}
  \caption{Statistics of human judgments collected in \benchmark{}.}
  \label{tab:labels}
\end{table}

\section{Experiments}
\label{sec:metrics}

To measure how well model preferences align with those of human participants from the conversation group, we define the following metrics and task setups.

\textbf{Ground truth of human preferences.}
For a candidate message judged by $N$ participants, each annotator either names a placement $(t_i,[s_i,e_i])$ or marks do-not-surface (DNS). Let $\mathcal{P}$ be the participants who place the message. The ground truth is the pair
$\mathcal{B} = \{\, t_i : i \in \mathcal{P} \,\}$ and $\mathcal{W} = \bigcup_{i \in \mathcal{P}} [s_i, e_i]$.
where $\mathcal{B}$ is the set of ideal turns they named, and $\mathcal{W}$ is the set of turns inside any of their acceptable windows. We define the ground truth as sets of turns instead of a single point in the conversation because of the inherent ambiguity of timing judgment. For a given message, appropriate turn-taking timing could vary with participants' knowledge, goals, and interpretation of the discussion~\cite{Cox2025SocialCM, Templeton2023LongGB}, so reducing these preferences to one turn would discard valid timing preferences. 

\textbf{Task setups.}
\label{sec:framings}
We evaluated 15 models from five providers, using standard and, where supported, maximum-reasoning configurations, for 28 configurations in total. Each configuration was evaluated across three task setups that vary in how much of the conversation is revealed before the model must commit, and consequently in how closely they reflect realistic deployment settings.

\begin{itemize}[nosep, leftmargin=1.2em, label={}, labelsep=0.4em]
  \item \textbf{(A) Turn-by-turn replay} (\textit{replay}): This is the most realistic deployment setting. The conversation is revealed one turn at a time, and the model decides at each turn whether to intervene. Once it answers ``yes,'' that turn becomes its final placement and cannot be revised.
  \item \textbf{(B) Replay with five-turn lookahead} (\textit{delayed replay}): This is a less constrained replay setting in which the model can see the next five turns at each yes-or-no decision.
  \item \textbf{(C) Selection from the full transcript} (\textit{pick}): This serves as an upper bound rather than a realistic deployment scenario. The agent sees the entire conversation and selects one turn.
\end{itemize}

\textbf{Metrics.}
We quantify performance across task setups using metrics defined against human timing preferences.

\begin{center}
\small
\begin{tabular}{l p{9.4cm}}
\toprule
\textbf{Metric} & \textbf{Definition} \\
\midrule
\textbf{Anyone's Ideal Turn} & The predicted turn is some annotator's ideal turn. \\
\textbf{Anyone's Window} & The predicted turn falls inside some annotator's window. \\
\textbf{Unanimous Window} & The predicted turn falls inside every annotator's window. Defined on 296 of 384 messages; NA elsewhere. \\
\textbf{Timing Drift} & Signed turns from the nearest ideal turn ($+$ late, $-$ early). \\
\textbf{Endorsement Share} & Fraction of participants who endorse the model's output as a ideal turn, an acceptable turn, or do-not-surface.  Endorsement share allows us to compare model performance against a human baseline: model share is
$\mathrm{endorsers}(x)/N$ while human share is
$(\mathrm{endorsers}(t_i)-1)/(N-1)$, the annotator's own vote removed from both
sides.\\
\bottomrule
\end{tabular}
\end{center}

\section{RQ1: How Do Participants Judge Good Timing?}
\label{sec:groundtruth}

\begin{figure}[t]
  \centering
  \includegraphics[width=\linewidth]{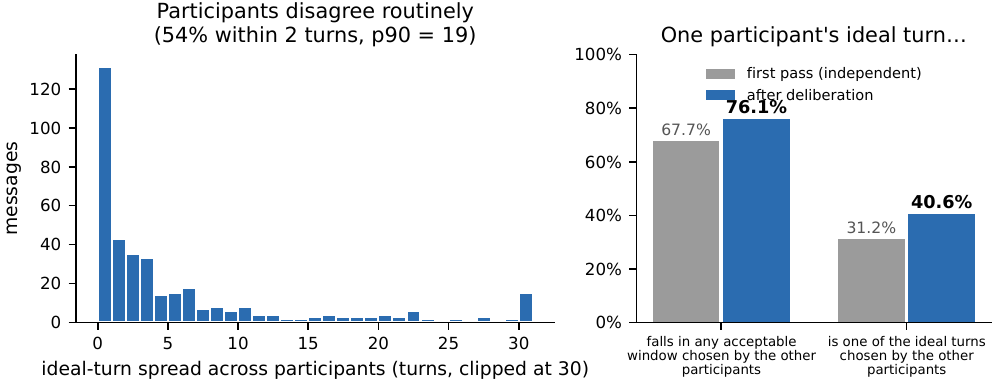}
  \vspace{-1.5em}
  \caption{Left: spread between participants' ideal turns. Right: participants scored against the rest
    of their own group. }
  \label{fig:agreement}
\end{figure}

\noindent\textbf{Participants agree on an acceptable window but disagree on ideal turn.}
\label{sec:regional}
An individual annotator’s acceptable window has a median width of 6 turns, covering 10.6\% of the corresponding conversation after normalizing each window by conversation length. When every annotator
shares at least one acceptable turn, the unanimous window has median width 3
turns. However, participants routinely disagree on the exact ideal turn placement (Figure~\ref{fig:agreement}). The
best-turn spread has median 2 turns and p90 19 turns, and only 30\% of messages
have a unanimous ideal turn. To further quantify the disagreement among annotations, we measure each annotator against the rest of their own group (i.e., \emph{endorsement share}). 
An annotator's chosen turn falls within others' acceptable windows 76.1\% of the time, but exactly matches their ideal turn only 40.6\%.

\noindent\textbf{Placements are spread through the conversation.}
Participants' placements of candidate messages were evenly spanned across the conversation. Across 1{,}047 placements, the mean relative position is 0.493, and the median is
0.486, with the middle half between 0.27 and 0.73. This broad distribution
indicates that there is no universally good time to place a candidate message. This distribution also
helps explain some annotator disagreement. A message can both reasonably reopen
a thread the group dropped or support a claim after the group returns to the
topic. We
therefore treat disagreement in annotation data as a
signal about the range of recipient preferences and define the model's target as a set of acceptable placements, 
rather than forcing the model to match a single averaged placement that may
reflect no annotator's judgment.

\section{RQ2: Can Prompted Models Match Human Timing preferences?}
\label{sec:results}
Models can often identify plausible placements when the full conversation is
visible, but the same models struggle when they must decide online (Table
\ref{tab:headline}).  

\begin{table}[t]
  \centering
  \small
  \begin{adjustbox}{max width=\linewidth,center}
  \begin{tabular}{l c c c c c}
    \toprule
    \textbf{Task Setup/Baseline} & \textbf{Anyone's} & \textbf{Anyone's} & \textbf{Unanimous} & \textbf{Median} & \textbf{Best} \\
     & \textbf{ideal turn} & \textbf{Window} & \textbf{Window}$^{\dagger}$ & \textbf{arm} & \textbf{arm} \\
    \midrule
    (A) turn-by-turn replay (\textit{replay})    & 19.0\% \tiny{[16.5, 21.7]} & 50.5\% \tiny{[46.8, 54.3]} & 24.2\% & 19.0\% & 32.8\% \\
    (B) replay with five-turn lookahead \emph{(delayed replay)} & 22.4\% \tiny{[19.3, 25.5]} & 52.2\% \tiny{[48.2, 56.1]} & 27.8\% & 22.1\% & 33.9\% \\
    (C) selection from the full transcript (\textit{pick})     & 57.8\% \tiny{[54.3, 61.2]} & 92.7\% \tiny{[90.9, 94.3]} & 73.3\% & 58.7\% & 64.8\% \\
    \midrule
    Human \textit{replay} baseline & 23.7\% & 53.7\% & 33.2\% & --- & --- \\
    \midrule
    Random turn policy                            & 3.7\%  & 25.5\% & 7.6\%  & --- & --- \\
    \bottomrule
  \end{tabular}
  \end{adjustbox}
  \hspace{-2em}
  \caption{Model rows are pooled over 28 configurations, with 95\% bootstrap
    intervals resampling items. Best arm is \texttt{gemini-3.6-flash} in both
\textit{replay} task setups and \texttt{gpt-5.6-sol} in \textit{pick}. 
    $^{\dagger}$Unanimous Window is defined on the 296 of 384 messages with a
    non-empty intersection across participants and is not comparable to the other
    columns. }
  \label{tab:headline}
\end{table}

\noindent\textbf{Models remain far below the participant-grounded target online.}
In \textit{replay}, model placements match some annotator's ideal turn on 19.0\% of
messages and fall inside some annotator's window on 50.5\%, leaving nearly half
of model placements outside every participant-endorsed window. Five-turn lookahead in the \textit{delayed replay} task setup
only marginally improves performance, reaching 22.4\% on Anyone's
ideal turn and 52.2\% on Anyone's Window. There is also considerable variance in accuracy 
 across configurations, from 6.2\% for \texttt{gpt-oss-120b} with
reasoning to 32.8\% for \texttt{gemini-3.6-flash} with reasoning, measured by ideal turn accuracy.
Full-conversation context substantially improves models' performance. With full hindsight in the \textit{pick} task setup, models' average accuracy rises from 50.5\% to 92.7\% on Anyone's Window, but still misses nearly half of the placements (57.8\%), judging by Anyone's ideal turn. 

\begin{figure}[t]
  \centering
  \includegraphics[width=0.85\linewidth]{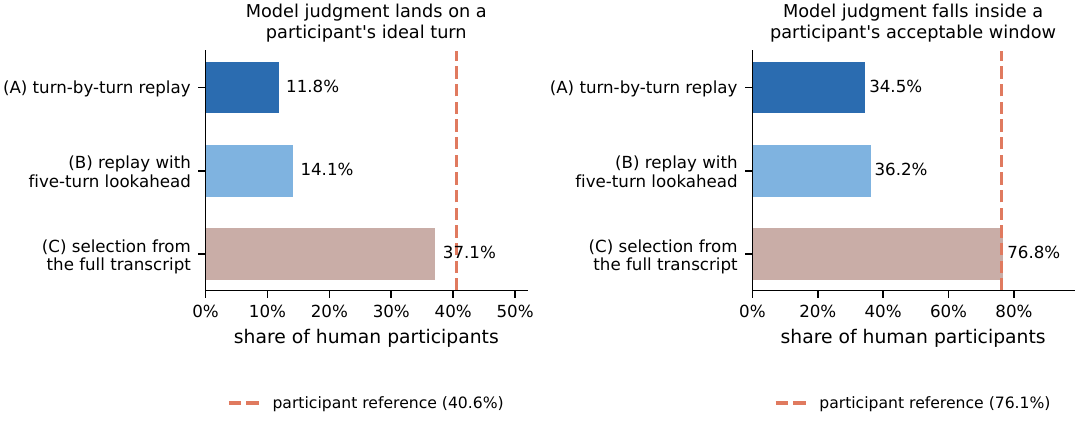}
   \vspace{-.5em}
  \caption{Share of the annotator group a single placement satisfies, against
    the human reference (dashed). \textit{Pick} reaches the human line; both \textit{replay}
    task setups fall far short of it.}
  \label{fig:endorsement}
\end{figure}

\noindent\textbf{Online models satisfy fewer in-group preferences than participants' own placements.} Endorsement share lets us compare humans and models using the same metric: the share of the original conversation group a single placement satisfies (Figure~\ref{fig:endorsement}).  
After deliberation, a participant's placement
on average satisfies 40.6\% of the rest of the group on the best-turn criterion and 76.1\%
on the window criterion; it satisfies the entire group 28.6\% and 63.2\% of the
time. Models reach participant-level endorsement only in the \textit{pick} task setup with hindsight of full conversation context (Appendix Table~\ref{tab:endorsement}). In \textit{replay}, models are 28.1 points below the deliberated
participant reference on best-turn share ([-32.4, -24.0]) and 40.6 points below it on window
share ([-44.4, -36.7]) (Appendix Figure~\ref{fig:endorsement-arms}). 
Under online evaluation, models achieve less than one-third of the human best-turn share and less than half of the window share.

\noindent\textbf{Models also underperform independent lay annotators in \textit{replay} mode.}
The original participants provide the benchmark reference but their
retrospective placements do not measure how people perform under the online constraint. We therefore ran a separate study with independent Prolific annotators to set a layperson baseline for the \textit{replay} task (study design is detailed in
Appendix~\ref{sec: human-baseline-study}). Of 81 recruited annotators, 56 annotators contributed 192 placements. 
We find lay annotators matched a participant's ideal turn on 23.7\% of placements and fell within a participant's union window on 53.7\% of placements; an improvement compared to 19.0\% and 50.5\% for models, respectively. 
Laypeople still fall short, suggesting the importance of both the additional context that participants bring and the inherent difficulty of anticipating where a discussion will head.

\begin{figure}[t]
  \centering
  \includegraphics[width=1.0\linewidth]{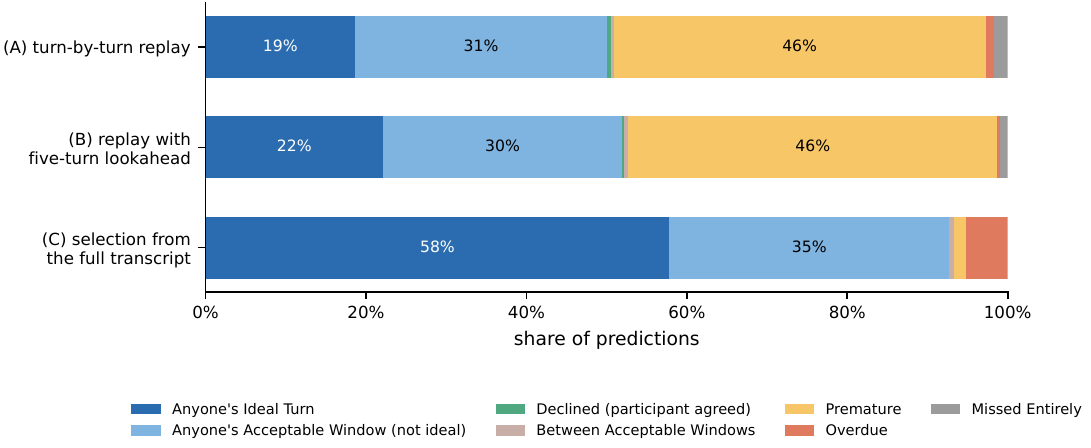}
  \vspace{-1.5em}
  \caption{Where predictions land. Online, the single dominant outcome is firing
    before any annotator's window opened; with hindsight it nearly vanishes and
    the residual error becomes lateness.}
  \label{fig:outcomes}
\end{figure}

\begin{figure}[t]
  \centering
  \includegraphics[width=1.0\linewidth]{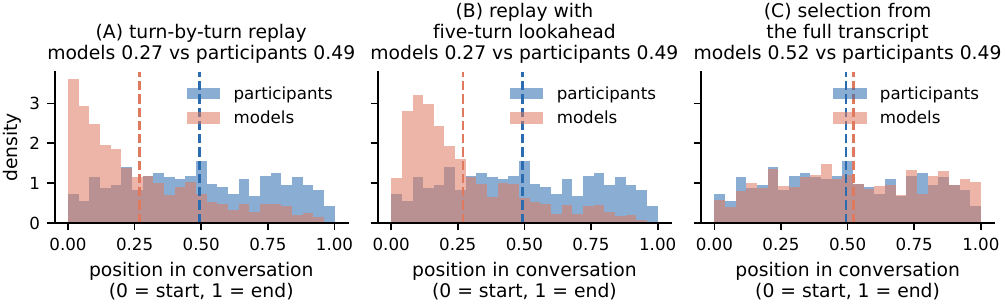}
\vspace{-1.5em}
  \caption{Where models place messages against where people do. The
    distributions barely overlap online but then nearly coincide with hindsight.}
  \label{fig:positions}
\end{figure}

\noindent\textbf{Errors in the online task setups are systematically premature.}
The dominant error in the online task setup is firing before any annotator's window starts (Figure~\ref{fig:outcomes}).
In \textit{replay}, 46.4\% of predictions are issued before
any annotator's window has opened, compared with 1.0\% issued after every window
has closed. This pattern indicates that models tend to place a given message before any of the acceptable windows have begun. In \textit{replay}, the average timing shift is $-11.1$ ($-15.2$ conditional on being early) (Figure~\ref{fig:drift} in Appendix). Participants
place messages on average 49.3\% of the way into a full conversation, while models chime in on average 26.9\% of the way (Figure~\ref{fig:positions}).
Considered against a median conversation length of 52 turns, 
this distribution shows that the average online prediction arrives
approximately one fifth of a conversation before the point at which the group
judged it appropriate.
Compared to models, we also find that lay annotators are less likely to chime in prematurely (39.7\% vs 46.4\%).

\noindent\textbf{Justifications at the point of commitment are most often about relevance, not timing.}
To examine what models attend to when they commit, we used a \texttt{deepseek-v4-pro} judge to classify 4{,}349 \textit{replay} trajectories from the
13 models with inspectable reasoning traces, withholding model identity and
prediction correctness (Appendix~\ref{sec:app-prompt-judge}).
In \textit{replay}, 92.8\% of
firing justifications argue that the content fits the current point in the
conversation, 7.0\% argue that the present moment is appropriate, and 0.14\%
explicitly weigh waiting. These labels suggest that when asked whether a message should be delivered at the current turn, models often answer whether the message is relevant now, rather than explicitly reasoning about whether the timing could be better in a later moment.

\section{Discussion, Limitations, and Future Work}
\label{sec:pitch-discussion}

\textbf{The same proactive message can be appropriate at multiple distinct points in a conversation.}
Half the messages in \benchmark{} have ideal placements more than two turns apart even after deliberation, and some cluster around multiple distinct points in a conversation. For example, a snack suggestion was placed when food was first raised and again when the topic returned much later. This motivates metrics and ground truth that preserve multiple valid timings rather than force a single gold turn. We found that participants tend to agree when a message has a clear conversational anchor, such as when a participant directly asks about the cost of a bus fare. They are more likely to disagree when the same message could be useful at multiple points in the conversation. For example, a comment about medical evaluation could either revive an earlier thread or support a related claim made twenty turns later. These cases suggest that some disagreements that remain after deliberation reflect different valid uses of the same message, rather than disagreement about the facts.

\textbf{What does the retrospective reference measure?}
\label{sec:app-fairness}
The reference in \benchmark{} is retrospective: participants see the full conversation and may revise placements after peers' choices, whereas replay models observe only prior turns. Alignment therefore measures how well models anticipate participants' timing preferences, rather than how they compare with humans making the same online decision. These retrospective preferences still provide a meaningful target for when recipients consider an intervention appropriate, but they cannot establish whether a live intervention would improve the discussion, since it may alter what follows.

Our data also captures preferences for only a single proactive interjection, and thus cannot model how prior interventions or intervention frequency shape later timing preferences. Future work could study these sequential dynamics through richer human data collection or simulation-based benchmarks as models become better at reproducing realistic group conversations.

Under \textit{replay}, model placements receive 34.5\% endorsement versus 76.1\% for participants' retrospective placements. Together with frequent premature interventions, this gap motivates policies that explicitly weigh acting now against waiting. Prompt ablations and an online human baseline could further separate model-specific limitations from decision framing and task difficulty. Similar challenges arise in proactive coaching, where whether, when, and how to intervene are distinct decisions~\cite{srinivas2025substance}.

\textbf{Current recruitment limitations and opportunities for broader participation.}
Our current participant pool reflects the practical demands of synchronous group sessions, with many participants recruited from university communities and through word of mouth. Future work could broaden the demographic range and examine how timing preferences relate to participants' roles and perceptions within a conversation, such as topic expertise, relative influence, or personality, as well as whether models align differently with different participants. The research discussions also center largely on AI, CS, HCI, and related computational fields; extending the protocol to natural sciences, humanities, and other domains could reveal additional conversational dynamics and timing patterns.

\textbf{Human baselines and the potential for ``superhuman'' proactive agents.} The considerable gap between retrospective participant judgments and the layperson online baseline suggests opportunities to go beyond matching online human decisions. With richer participant context and better models of how the conversation may unfold, an agent could potentially anticipate preferred intervention points more accurately than a layperson deciding in real time. Another direction is to explicitly trade off placement quality against latency by allowing the agent to wait briefly before intervening. In threaded text chat, interface design allows for this tradeoff by enabling an agent to insert a contribution into the recent conversational history, making slightly delayed but better-placed interventions feasible. Indeed, our \textit{delayed replay} task setup probes this possibility, showing some limited improvements in placement accuracy with up to a 5-turn delay in timing.

\section{Related Work}
\label{sec:pitch-related}

\paragraph{Proactive AI agents.} The question of when an agent should act dates back to mixed-initiative systems, where assistance must balance the value of acting against the cost of interruption~\citep{horvitz1999principles}. Recent proactive agents increasingly model timing explicitly, often with a gate that decides whether the agent should speak~\citep{mao2024muca, eckhaus2025time, chen2025llamapie, yang2026proact, huang2026chatagents}. Related benchmarks evaluate this through per-turn \textsc{Speak}/\textsc{Silent} prediction~\citep{patel2025discussllm, chen2025llamapie, nama2026when2speak}, but typically designate a single correct intervention point, often in synthetic or semi-synthetic conversations. In contrast, \benchmark{} uses unscripted human conversations and timing judgments from the actual participants, allowing multiple acceptable placements and treating persistent disagreement as meaningful variation~\citep{aroyo2015truth, plank2022problem, abdulhai2025consistently}.

\paragraph{Human--AI collaborative agents.}
A parallel body of work studies AI agents in human conversations and collaborative group settings~\citep{do2022how, houde2025controlling, yeo2026group}. In groups, proactive timing is especially consequential because an intervention enters a shared space and draws multiple participants' attention~\citep{yeo2026group}. Prior work often evaluates downstream outcomes or predefined intervention policies rather than whether timing aligns with participants' preferences. \benchmark{} complements this literature by isolating the timing of a fixed contribution and evaluating it against participant-grounded judgments.

\section{Conclusion}

We introduce \benchmark{}, a benchmark for evaluating when proactive agents should intervene in multi-party conversations. By jointly collecting unscripted human conversations, participant-authored messages, and participants' timing judgments, \benchmark{} provides a human-grounded testbed for studying proactive timing. Our results show that current models often intervene too early and that appropriate timing can vary across participants and conversational contexts. We hope \benchmark{} enables future work on agents that better understand not only what to contribute, but when to contribute it.

\newpage
\clearpage

\subsection*{Ethics statement}

The study was reviewed and approved by our institution's IRB. Participants gave
informed consent to the public release of the conversation they held and the
annotations they produced, were compensated \$35 for the session, and could
withdraw at any point, including after the fact. Participants' names are removed from the released dataset. Due to constraints in the data collection procedures, the corpus has limitations: it is
English-language text among small, broadly egalitarian groups drawn largely from
research and social networks in the United States, so a system calibrated to it
encodes one set of conversational norms.

\subsection*{Reproducibility statement}

Section~\ref{sec:study} gives the collection protocol.
Section~\ref{sec:metrics} defines every metric.
Section~\ref{sec:framings} specifies the three task setups and
the decoding settings; Appendix~\ref{sec:app-prompts} gives the prompts. Appendix~\ref{sec:app-models} lists the 15 models and 28
configurations together with coverage and cost. Appendix~\ref{sec:app-models}
gives the construction of the no-model reference random policy. The dataset and the analysis notebooks, and the figure generation script are included in the supplementary material. We will release the code of the annotation systems upon acceptance of the paper. 

\subsection*{AI use statement}
We used generative AI as part of the research methodology to classify model reasoning traces, as described in Section~\ref{sec:results}. Second, we used AI tools to assist with paper revision and language editing. Third, we used coding agents to help develop the data collection interface and supporting system, and to assist in writing data analysis code. The authors reviewed and verified all AI-assisted writing and code, and the reported analyses were produced by running the verified analysis code. Generative AI was not used to make final research conclusions or replace author review and interpretation. Other uses of generative AI were not applicable.

\clearpage
\newpage

\bibliography{references}
\bibliographystyle{iclr2027_conference}

\newpage
\clearpage

\appendix

\section{Conversation Topics}
\label{sec:convo-prompts}
The benchmark evaluates knowledge placement within ongoing conversations spanning a diverse set of topics and conversational goals. Each conversation begins with a prompt developed from topic interests the users provided in a pre-study survey. These are designed to cover a range of domains and information needs.

Below, we provide two representative prompts in full, followed by a list of keywords describing the broader prompt set to illustrated the breadth of topics and activities represented in the benchmark. Candidate knowledge messages associated with these conversations are described separately in Appendix~\ref{sec:candy-msg}.

\subsection{Representative Topics}

\paragraph{Collaborative Planning.}

\textit{``You're planning a 3-day trip together to a US city for Labor Day weekend. Plan their itinerary together, a mix of must-see spots and a couple of local, less touristy picks, plus where you should eat and how to get around, given the time you have. Consider the hours, current prices, transportation, etc. By the end, land on: a city, a day-by-day itinerary, and a rough per-person budget.''}

This prompt represents the collaborative planning subset of the benchmark, in which participants may develop travel itineraries or event plans together and encounter information needs over the course of the conversation.

\paragraph{Scholarly Discussion.}
\textit{``What Does HCI Owe AI Development Right Now? The dominant framing in AI research is capabilities: can the model do the task? HCI has decades of knowledge about interaction patterns, user agency, and sociotechnical systems, but is that knowledge actually reaching the people building agents? What concrete contributions should HCI researchers be pushing into AI development today, and what's stopping them? Is formalization the bottleneck, or is it access and incentives?''}

This prompt represents the scholarly discussion subset, where the conversation develops around an open-ended research question rather than a concrete planning task.

\subsection{Topic Breadth}

Across the benchmark, 25 prompts span a range of travel, event organization, learning, and budget-constrained planning as well as AI in education and research, evaluation, design, and accessibility.

\textbf{Collaborative planning}
\begin{itemize}
    \item Plan a two- to three-day trip
    \item Plan a reunion weekend
    \item Organize a small creative hackathon
    \item Organize an esports competition
    \item Plan a lab group retreat
    \item Learn a skill together
    \item Run a budget-constrained event
\end{itemize}
\textbf{Research discussions}
\begin{itemize}
    \item Explore AI tools for student learning
    \item Consider AI for research assistance
    \item Discuss existing AI benchmarks and progress measurement
    \item Examine generative AI for user simulation
    \item Consider AI beyond the productivity paradigm
    \item Discuss who should determine AI agent behavior control
    \item Examine conversational AI voice and tone
    \item Consider LLM agents for accessibility
    \item Discuss LLM assumptions across diverse users
    \item Explore unresolved issues around autonomous driving technology
    \item Examine how HCI knowledge can inform AI development
\end{itemize}

\section{Candidate Knowledge Messages}
\label{sec:candy-msg}

For each conversation, participants produce candidate knowledge messages that they believe appropriate to introduce at a particular point in their dialogue. We provide five representative candidate messages below, illustrating the range of knowledge being evaluated.

\textbf{Example 1:}

``If you're looking for more budget-friendly seafood options beyond the upscale spots, there are some great affordable alternatives. [Yankee Lobster Co.](\url{https://www.busybeingshasha.com/blog/yankeelobster}) gets consistently recommended for reasonably priced lobster rolls, and [Pauli's](\url{https://www.tripadvisor.com/Restaurants-g60745-c33-zfp16-Boston\_Massachusetts.html}) shows up on multiple "best cheap seafood" lists. For a true local experience, check out Burke's - it's described as the definition of "hole in the wall" seafood on [r/boston](\url{https://www.reddit.com/r/boston/comments/14kyu1q/best\_hole\_in\_the\_wall\_cheap\_seafood}), praised for being professional and clean but not fancy. You could definitely do a good seafood meal for much less than the ``splurge'' places while still getting that authentic Boston experience.''

This example corresponds to the trip-planning prompt in Appendix~\ref{sec:convo-prompts}.

\textbf{Example 2:}

AI researchers may be undervaluing interface design, a concern that is receiving attention from design leaders. Jakob Nielsen recently argued we might be heading toward ["No More User Interface"](\url{https://jakobnielsenphd.substack.com/p/no-more-ui}) as AI becomes the primary interaction layer. However, there's a counter-argument that [AI is exposing what design really is](\url{https://uxdesign.cc/what-ai-exposes-about-design-319029d48441}) - the strategic thinking behind interfaces, not just the visual output. Some companies are replacing UX research with [AI-generated user testing](\url{https://smart-interface-design-patterns.com/articles/ai-user-research}), which could further marginalize HCI contributions if the field doesn't articulate its unique value in understanding human-AI interaction patterns.

This example corresponds to the scholarly-discussion prompt in Appendix~\ref{sec:convo-prompts}.

\textbf{Example 3:}

That's a really important point about dialects. Research shows that reward models used in LLM training exhibit systematic biases against non-standard dialects like African American Language [**Rejected Dialects: Biases Against African American Language in Reward Models**](\url{https://www.semanticscholar.org/paper/f400b0d5724203633f56299cb19f34bf18fd31b9}). So longer context windows might not help if the underlying training biases cause models to misinterpret dialectal variations.

\textbf{Example 4:}

Boston's on the Northeast Corridor, which is actually Amtrak's strongest route! The Northeast Regional runs frequently between Boston and NYC(\url{https://www.timeout.com/usa/news/a-defense-of-the-northeast-regional-amtraks-best-east-coast-train-111424}) (about 3.5 hours) and continues down to DC. It's much more reliable than most west coast Amtrak routes. Plus you could potentially take the Acela for the premium experience - that gets you to NYC in under 3 hours.

\section{Study Platform for the Interactive data collection protocol}
\label{sec:interface-protocol}

We developed a web-based data collection platform to collect conversations while supporting both candidate knowledge message creation and cross-annotation. It hosts the full study workflow.

\begin{figure}[h]
  \centering
  \includegraphics[width=\linewidth]{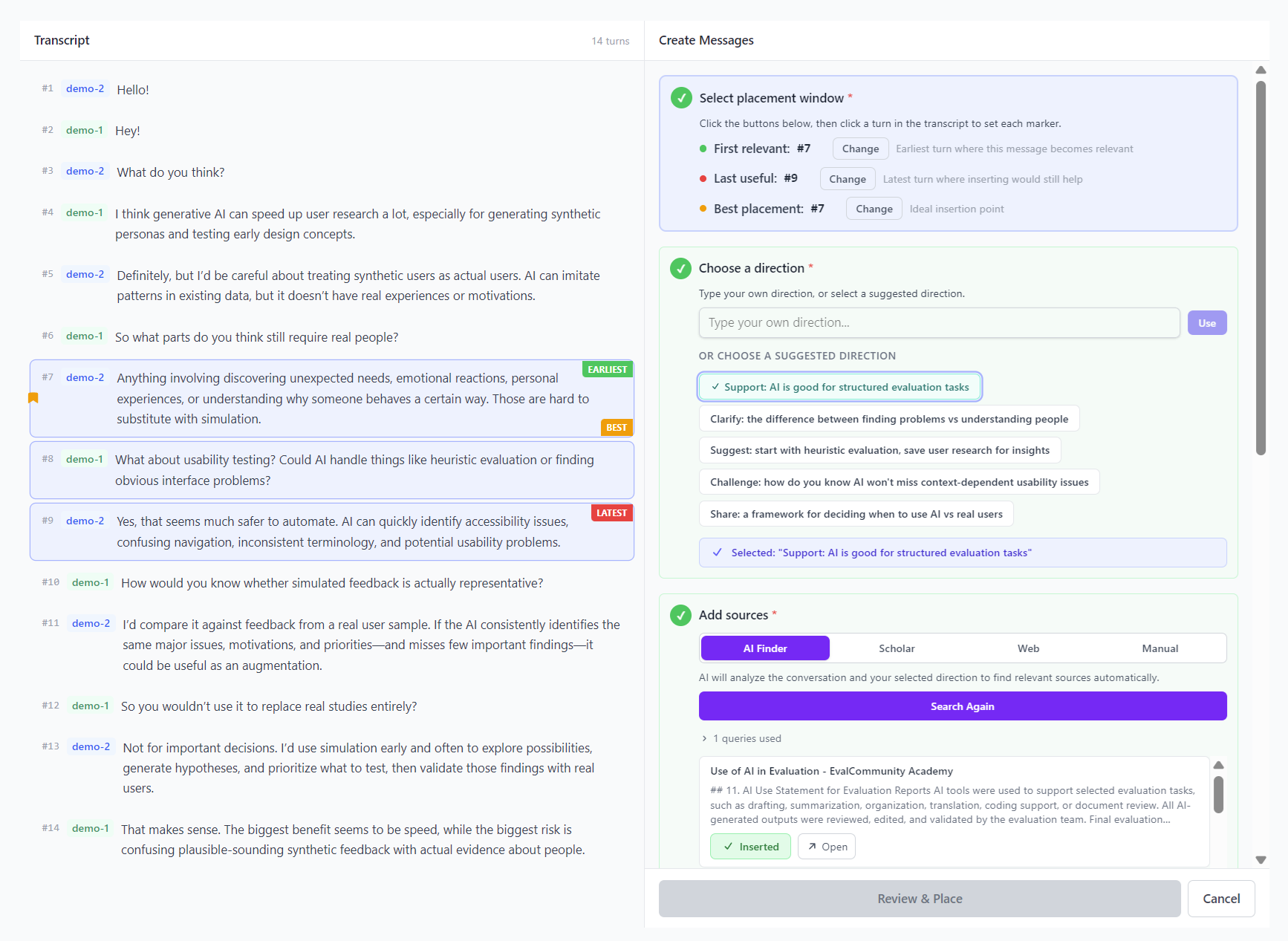}
  \caption{Message contribution interface. The full transcript sits in a scrollable
    pane on the left and the candidate message and placement controls on the
    right. }
  \label{fig:interface-app}
\end{figure}

\paragraph{Conversation and contribution.}
Participants first use the platform to engage in a synchronous group conversation. After the conversation, they review the transcript and create candidate knowledge items they believe could have been useful during the discussion. The message contribution interface (Figure~\ref{fig:interface-app}) requires participants to identify where in the conversation the item would be relevant, including ideal turn and a range of acceptable turns. Integrated tools help with resource search and retrieval as well as message authoring.

\begin{figure}[h]
  \centering
  \includegraphics[width=\linewidth]{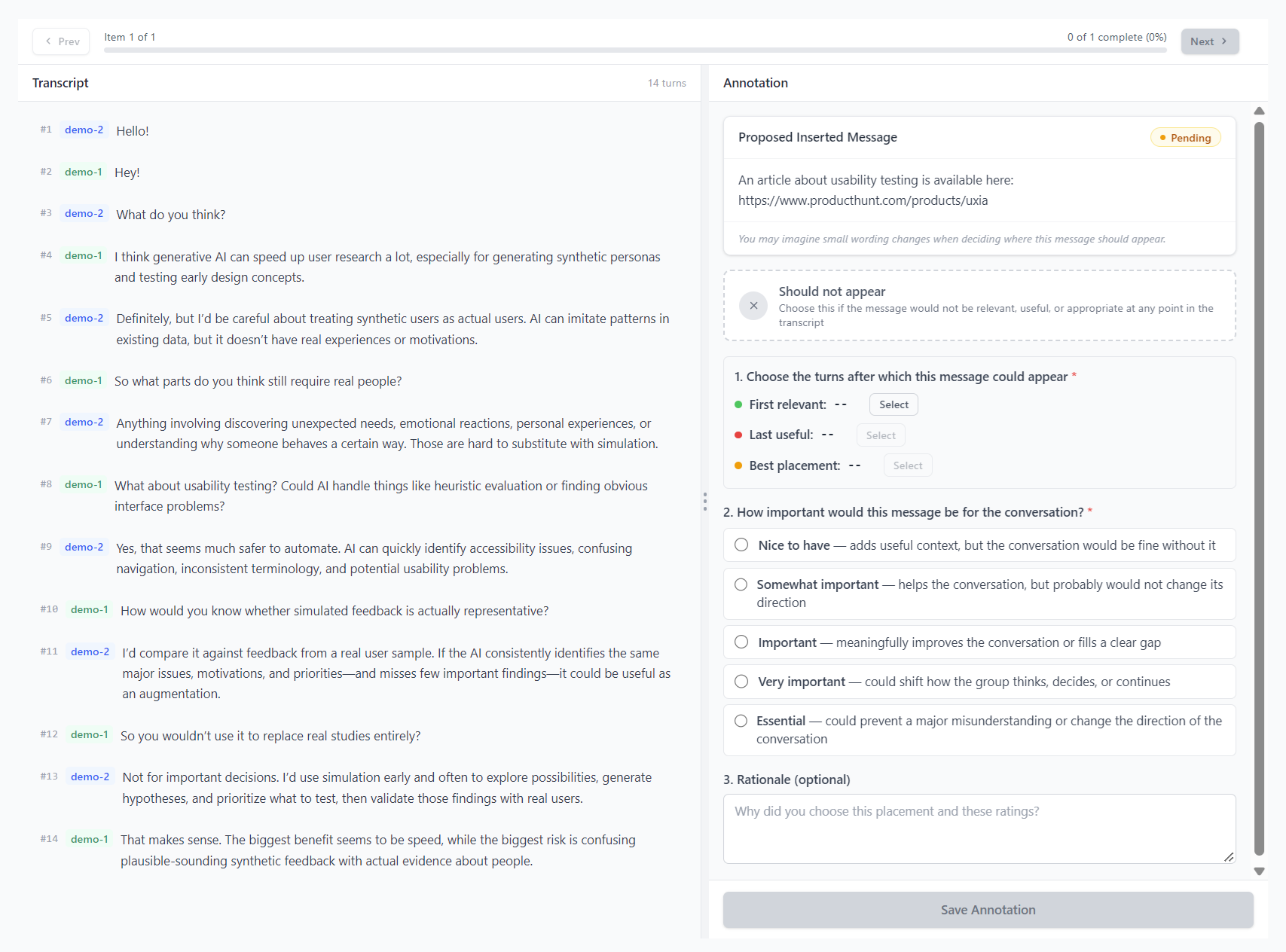}
  \caption{Timing judgment interface. Users navigate the transcript in the left pane while making annotation decisions in the right pane for each message contributed by other participants in the group.}
  \label{fig:anno-interface-app}
\end{figure}

\paragraph{Timing judgments.} Post-construction, participants play the role of participants, reviewing the neutralized (Appendix~\ref{sec:app-prompt-neutral}) contributed messages created by each other. The interface allows them to similarly indicate where these should appear within the conversation, if at all, and optionally providing brief rationale for their decisions (Figure~\ref{fig:anno-interface-app}).

\begin{figure}[h]
  \centering
  \includegraphics[width=\linewidth]{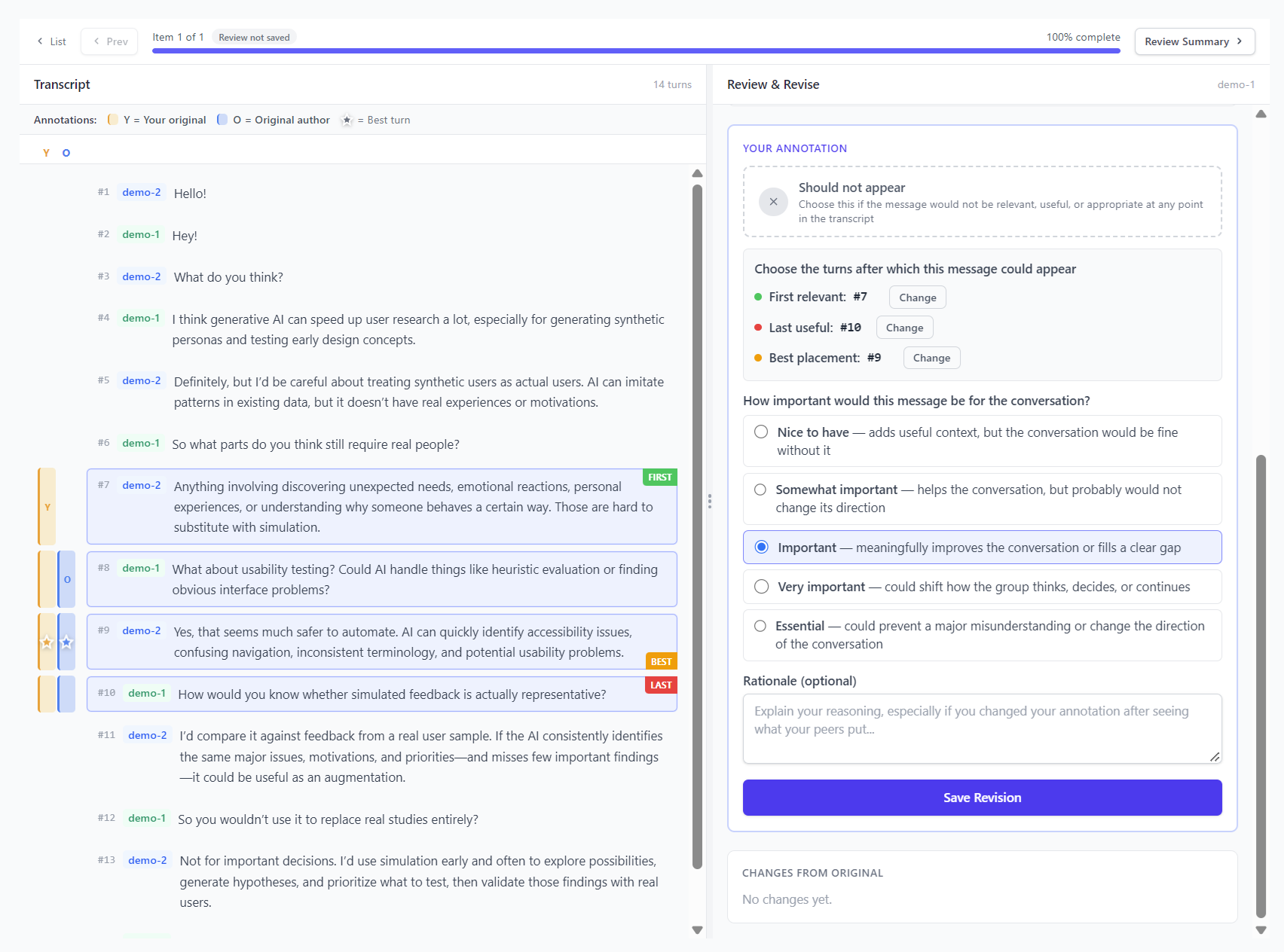}
  \caption{Deliberation interface. Users are presented with the conversation transcript once again, alongside timing judgments made on a contribution message by all other participants. The right pane allows users to optionally revise and update their judgments.}
  \label{fig:deliberation-app}
\end{figure}

\paragraph{Deliberation.} A subsequent deliberation stage presents each item from the set of created knowledge messages with every participant's placement, allowing participants to review and revise their own annotations (Figure~\ref{fig:deliberation-app}).

Researchers manage study group formation, review knowledge items, configure annotation assignments, and export the resulting data through an admin interface. The platform also includes an interactive tutorial to familiarize users with each stage of the study before they begin the actual task.

The platform was implemented as a full-stack web application using React and TypeScript with a FastAPI PostgreSQL backend.

\section{Human Baseline for Intervention Timing}
\label{sec: human-baseline-study}

To contextualize model performance, we conducted a human baseline study measuring how lay users perform the same replay intervention-timing task. The goal was to determine whether models' difficulty in selecting an appropriate intervention point and tendency to speak early reflects the difficulty of the task or a gap between model behavior and human preferences.

\subsection{Study Design}

Participants were recruited through Prolific and completed the replay task (Section~\ref{sec:framings}) on a web-based annotation platform. We collected data across both scholar and trip conversation scenarios which allows us to measure human baseline performance across the full range of contexts used in model evaluation.

Our ground truth is derived from the data collected for PITCH. For each candidate contribution, we considered two measures: ideal turn (the set of turns selected as ideal intervention points in our ground truth) and union windows (span of earliest to latest acceptable intervention points across participants). These measures allow direct comparison with model performance.

\subsection{Annotation Platform and Protocol}

The annotation interface presented participants with a conversation transcript alongside three corresponding candidate contributions, de-anonymized and pulled from the PITCH dataset. The interface simulates a natural flow of conversation by allowing participants to progress turn by turn, deciding at each point whether any of the three candidate contributions should be sent. This reflects the replay task we used for model evaluation.

At each turn in the conversation, participants could see:
\begin{itemize}
    \item The conversation history up to that point
    \item The three candidate contributions
    \item A ``Send'' button for each contribution that had not yet been placed
    \item A ``Continue'' button to advance to the next turn without placing any contribution
\end{itemize}

If participants reached the end of the conversation with contributions still unplaced, they were given the option to mark those remaining contributions as ``not relevant''--indicating the message did not fit anywhere in the conversation.

Each session consisted of placing three candidate contributions within a single conversation and participants could complete multiple sessions across different conversations.

Participants received a base compensation of \$1.00 USD, with up to an additional \$2.00 bonus for placing contributions within the ground truth union windows. This incentive structure encouraged accuracy while acknowledging that reasonable people may disagree on exact timing.

\subsection{Participants and Quality Control}

We recruited 109 participants, of which 59 completed selections for collaborative-planning scenarios and 50 for scholarly-discussion scenarios. After manual quality control, 331 data points were retained for analysis.

Quality control consisted of filtering for suspicious completion patterns: users who completed items in implausibly short times or who selected the first available turns regardless of conversation content suggested that they did not read the instructions or conversations carefully.

\subsection{Results}

The resulting lay-user baseline substantially exceeds the performance of models evaluated on the same task (Table~\ref{tab:app-human-baseline}).

\begin{table}[t]
    \centering
    \caption{Lay-user and model performance on intervention timing. Human performance is reported separately for scholarly and collaborative scenarios, as well as in aggregate (N=331 placements).}
    \label{tab:app-human-baseline}
    \begin{tabular}{lccc c}
        \toprule
        & \multicolumn{3}{c}{\textbf{Human (Lay Users)}} & \textbf{Models} \\
        \cmidrule(lr){2-4}
        \textbf{Metric} & \textbf{Scholarly} & \textbf{Collaborative} & \textbf{Aggregate} & \\
        \midrule
        Anyone's ideal turn & 22.4\% & 24.6\% & 23.7\% & 19.0\% \\
        Within anyone's window & 55.2\% & 52.6\% & 53.7\% & 50.5\% \\
        Within unanimous window & 34.3\% & 32.5\% & 33.2\% & 24.2\% \\
        Earlier than anyone's window & 37.6\% & 41.1\% & 39.7\% & 46.4\% \\
        Mean timing shift (turns) & $-4.3$ & $-11.8$ & $-8.7$ & $-11.1$ \\
        \bottomrule
    \end{tabular}
\end{table}

Both lay users and models exhibit a consistent early placement bias. Participants placed contributions earlier than anyone's window in 39.7\% of cases, with a mean timing shift of $-8.69$ turns from the nearest acceptable intervention point. However, this bias is notably less severe than models' tendency to intervene early.

\section{Models, Coverage, and Cost}
\label{sec:app-models}

\begin{table}[t]
  \centering
  \small
  \begin{adjustbox}{max width=\linewidth,center}
\begin{tabular}{l l}
    \toprule
    \textbf{Provider} & \textbf{Models} \\
    \midrule
    Google       & \texttt{gemini-3.1-pro}, \texttt{gemini-3.6-flash} \\
    OpenAI       & \texttt{gpt-5.6-sol}, \texttt{gpt-5.6-luna} \\
    Anthropic    & \texttt{opus-4-6}, \texttt{sonnet-4-6}, \texttt{haiku-4-5} \\
    \addlinespace
    \multirow{3}{*}{Open-weight (Together AI)}
                 & \texttt{deepseek-v4-pro}, \texttt{glm-5.2}, \texttt{kimi-k3}, \texttt{minimax-m3} \\
                 & \texttt{qwen-3.7-max}, \texttt{gpt-oss-120b}, \texttt{inkling} \\
                 & \texttt{nemotron-3-ultra-550b-a55b} \\
    \bottomrule
  \end{tabular}
\end{adjustbox}
  \caption{The 15 models evaluated, each in standard and (where supported) reasoning mode.}
  \label{tab:models}
\end{table}
\paragraph{Coverage and input.}
Models receive the neutralized candidate message and a system prompt describing
the proactive-agent task, with conversation visibility determined by the framing
(Section~\ref{sec:framings}). The prompt is included in Appendix~\ref{sec:app-prompts}. Standard mode uses temperature 1.

\paragraph{Scale and cost.}
The run comprises 77{,}755 API calls at a total cost of \$507.83: \$262.89 for replay against \$18.05 for pick.

\paragraph{Random Baseline} A random turn drawn uniformly over the
conversation.

\section{Additional Results}
\label{sec:app-results}

\begin{figure}[t]
  \centering
  \includegraphics[width=1.0\linewidth]{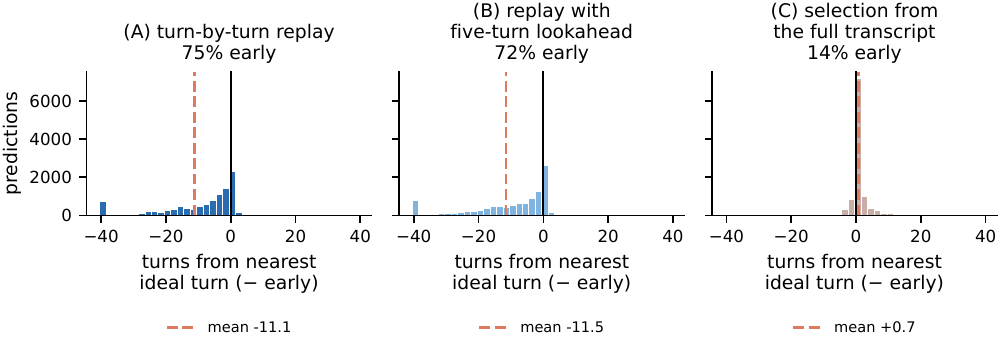}
  \caption{Timing drift by framing. Averaging the online and hindsight
    distributions would report a mild early bias that describes neither, which
    is why the framings are never pooled.}
  \label{fig:drift}
\end{figure}

\begin{table}[t]
  \centering
  \small
  \begin{adjustbox}{max width=\linewidth,center}
  \begin{tabular}{l c c c c}
    \toprule
     & \multicolumn{2}{c}{\textbf{Best-turn criterion}} & \multicolumn{2}{c}{\textbf{Window criterion}} \\
    \cmidrule(lr){2-3}\cmidrule(lr){4-5}
    & mean share & satisfied all & mean share & satisfied all \\
    \midrule
    \textbf{Human annotator, after deliberation} & 40.6\% & 28.6\% & 76.1\% & 63.2\% \\
    \textbf{Human annotator, first pass}         & 31.3\% & 18.0\% & 67.7\% & 53.2\% \\
    \midrule
    \textit{(A) turn-by-turn replay}     & 37.1\% & 18.1\% & 76.8\% & 56.5\% \\
  \textit{(B) replay with five-turn lookahead} & 14.1\% & 6.7\%  & 36.2\% & 21.4\% \\
    \textit{(C) selection from the full transcript}    & 11.8\% & 5.4\%  & 34.5\% & 18.7\% \\
    \bottomrule
  \end{tabular}
  \end{adjustbox}
  \caption{Share of the annotator group a single placement satisfies, the only
    quantity on which humans and models are scored by an identical definition.
    Bold marks the two human reference rows. Both human rows exclude the
    annotator's own vote from numerator and denominator; a model has no vote in
    the group, which is what makes the rows comparable. The first-pass row is
    each annotator's independent judgment before seeing anyone else's. Paired
    differences reported in the text are computed on the \textit{n} items where
    every arm produced a placement, so they will not equal differences of these
    marginal columns.}
  \label{tab:endorsement}
\end{table}

\begin{figure}[t]
  \centering
  \includegraphics[width=1.0\linewidth]{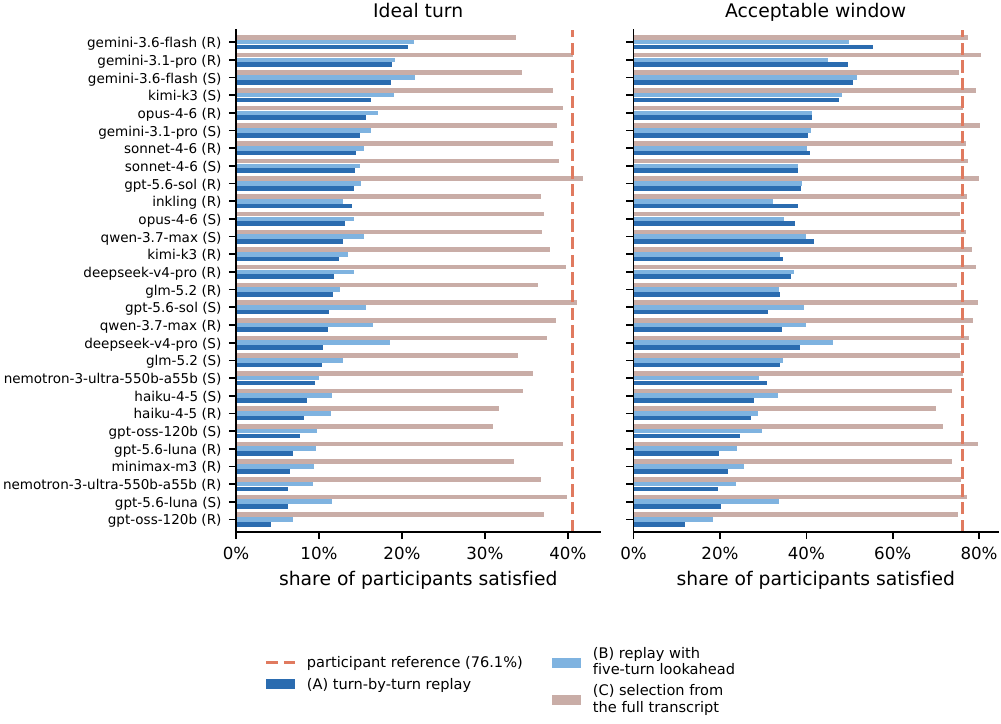}
  \caption{Every configuration against the human reference (dashed). In pick,
    three arms clear the human best-turn bar and eighteen clear the window bar;
    in both online framings, none do. (R) means reasoning mode and (S) means standard model configuration}
  \label{fig:endorsement-arms}
\end{figure}

\subsection{Model Performance by Conversation Scenario}
\label{sec:app-scenario-results}

Table~\ref{tab:scenario-results} separates the full-corpus model results in
Table~\ref{tab:headline} into scholarly discussions and trip-planning
conversations. The scenario split is smaller than the difference among task
framings, and its direction changes with the framing. In replay, trip-planning
accuracy is 2.0 points higher on Anyone's ideal turn, while Anyone's Window is
nearly unchanged (50.8\% against 50.3\%). With five-turn lookahead, scholarly
discussions are 2.5 points higher on Anyone's ideal turn. The largest scenario
gap appears in pick: scholarly discussions are 4.9 points higher on Anyone's
ideal turn, 3.4 points higher on Anyone's Window, and 5.8 points higher on
Unanimous Window.

\begin{table}[t]
  \centering
  \small
  \begin{adjustbox}{max width=\linewidth,center}
  \begin{tabular}{l l c c c c c}
    \toprule
    \textbf{Task Setups} & \textbf{Subset}
      & \textbf{Anyone's} & \textbf{Anyone's} & \textbf{Unanimous}
      & \textbf{Median} & \textbf{Best} \\
    & & \textbf{ideal turn} & \textbf{Window}
      & \textbf{Window}$^{\dagger}$ & \textbf{arm} & \textbf{arm} \\
    \midrule
    \textit{(A) turn-by-turn replay}  
      & Full
      & 19.0\% \tiny{[16.5, 21.7]}
      & 50.5\% \tiny{[46.8, 54.3]}
      & 24.2\% & 19.0\% & 32.8\% \\
    & Scholar
      & 18.2\% \tiny{[14.8, 21.9]}
      & 50.3\% \tiny{[45.6, 55.0]}
      & 23.6\% & 18.1\% & 28.3\% \\
    & Trip
      & 20.2\% \tiny{[16.2, 24.2]}
      & 50.8\% \tiny{[45.0, 56.2]}
      & 25.0\% & 20.6\% & 39.9\% \\
    \midrule
    \textit{(B) replay with five-turn lookahead} \emph{(eased)}
      & Full
      & 22.4\% \tiny{[19.3, 25.5]}
      & 52.2\% \tiny{[48.2, 56.1]}
      & 27.8\% & 22.1\% & 33.9\% \\
    & Scholar
      & 23.4\% \tiny{[19.3, 27.6]}
      & 53.7\% \tiny{[48.3, 58.8]}
      & 30.2\% & 24.6\% & 32.3\% \\
    & Trip
      & 20.9\% \tiny{[16.6, 25.4]}
      & 50.1\% \tiny{[44.2, 55.9]}
      & 24.9\% & 21.5\% & 37.3\% \\
    \midrule
    \textit{(C) selection from the full transcript} \emph{(upper bound)}
      & Full
      & 57.8\% \tiny{[54.3, 61.2]}
      & 92.7\% \tiny{[90.9, 94.3]}
      & 73.3\% & 58.7\% & 64.8\% \\
    & Scholar
      & 59.8\% \tiny{[54.9, 64.3]}
      & 94.1\% \tiny{[92.3, 95.8]}
      & 75.9\% & 60.2\% & 66.8\% \\
    & Trip
      & 54.9\% \tiny{[49.8, 60.0]}
      & 90.7\% \tiny{[87.4, 93.6]}
      & 70.1\% & 55.7\% & 63.3\% \\
    \bottomrule
  \end{tabular}
  \end{adjustbox}
  \caption{Model performance on the full corpus and by conversation scenario,
    pooled over 28 configurations. Full contains 384 messages, Scholar 226,
    and Trip 158. Brackets are 95\% bootstrap intervals resampling items.
    The best configuration differs across subsets:
    \texttt{gemini-3.6-flash} is best in replay, except that its standard arm
    leads Scholar and its reasoning arm leads Full and Trip;
    \texttt{kimi-k3} standard leads Scholar in delayed replay, while
    \texttt{gemini-3.6-flash} reasoning leads Full and Trip; and
    \texttt{gpt-5.6-sol} reasoning leads Full and Scholar in pick, while its
    standard arm leads Trip.
    $^{\dagger}$Unanimous Window is defined on 296 of 384 Full, 160 of 226
    Scholar, and 136 of 158 Trip messages.}
  \label{tab:scenario-results}
\end{table}

These are descriptive subgroup results rather than a controlled scenario
effect. Scenario is omitted from the model prompt, but the two subsets differ
in both conversation structure and message composition. Scholarly discussions
have a median of 42 turns and three participants, whereas trip-planning
conversations have a median of 79.5 turns and two participants
(Table~\ref{tab:conversations}). The bootstrap intervals for Scholar and Trip
also overlap in every framing. The split therefore does not support attributing
the observed differences to the scenario label itself.

\begin{figure}[h]
  \centering
  \includegraphics[width=0.95\linewidth]{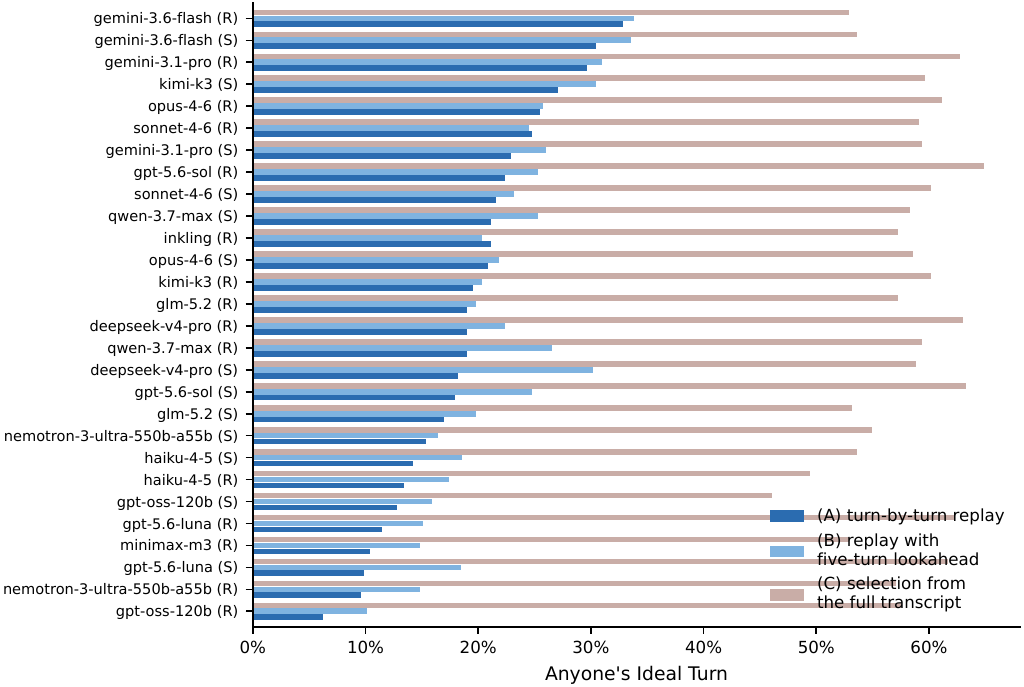}
  \caption{Anyone's ideal turn for all 28 configurations. The gap between the
    pick bar and the replay bar is wider than the entire spread across models
    within either framing, and it is positive for every configuration. (R) indicates models in native reasoning mode with max effort; (S) indicates models in lowest reasoning effort.}
  \label{fig:headline}
\end{figure}

\section{Prompts}
\label{sec:app-prompts}

The strings below are the prompts used to draft candidate messages, strip placement cues from them, elicit timing decisions, and label firing justifications. A token in braces is a runtime field. Every other character is fixed.

\subsection{Timing}
\label{sec:app-prompt-timing}

The same system message and user-message scaffold are used for pick, replay, and delayed replay. The reported standard and native-reasoning configurations both instruct the model to leave the reasoning field empty; native reasoning, where a provider supports it, is the configuration marked (R) in Figure~\ref{fig:headline}.

\paragraph{System message.}
\begin{promptblock}
You are observing a naturally unfolding small-group conversation. Conversations typically span roughly 20–200 turns, averaging about 60. You hold information that could help the group's conversation become more informative, productive, and smooth, which you would contribute in the form of a message inserted into the conversation.
\end{promptblock}

\paragraph{User-message scaffold.}
\begin{promptblock}
Visible conversation:
{formatted visible transcript}

Candidate messages to insert (each message appears once):
{candidate_messages_json}

{task_instruction}

{batch_instruction_if_any}

{reasoning_instruction}

Return exactly one valid JSON object matching this shape:
{answer_template_json}

{yes_example_if_replay_or_delayed_replay}

Do not use Markdown fences or include any text outside the JSON object.
\end{promptblock}

Each visible turn is written as below, with a blank line between turns:
\begin{promptblock}
Turn {turn_number} ({speaker}): {text}
\end{promptblock}

Candidate messages use request-local labels, not UUIDs:
\begin{promptblock}
[
  {
    "message_id": "item_1",
    "message": "{neutralized candidate message text}"
  },
  {
    "message_id": "item_2",
    "message": "{neutralized candidate message text}"
  }
]
\end{promptblock}

\paragraph{Task instruction, (A) turn-by-turn replay.}
\begin{promptblock}
At checkpoint {checkpoint}, decide whether to insert each message now, immediately after turn {checkpoint}. Use decision "YES" or "NO"; YES must select turn {checkpoint}, and NO must set selected_turn to null.
\end{promptblock}

\paragraph{Task instruction, (B) replay with five-turn lookahead}
\begin{promptblock}
You are now at turn {visible}. Decide whether to insert each message at one of turns {start} through {end} - the {span} most recent turns. The message would be inserted immediately after the turn you select. Use decision "YES" or "NO"; YES must select a turn between {start} and {end}, and NO must set selected_turn to null.
\end{promptblock}

\paragraph{Task instruction, (C) selection from the full transcript .}
\begin{promptblock}
For each message, select the single ideal turn in the full conversation to insert it. The message would be inserted immediately after the turn you select. Use decision "PICK" and its selected_turn.
\end{promptblock}

\paragraph{Batched-message instruction.}
Inserted for \texttt{\{batch\_instruction\_if\_any\}} when a request contains more than one candidate:
\begin{promptblock}
Judge each message completely independently, as if it were the only message you were given. Do not rank or compare the messages against one another, and do not let your decision about one message influence your decision about any other. Two messages may be inserted at the same turn, and any number of them may be inserted or not inserted.
\end{promptblock}

\paragraph{Answer template, (A) turn-by-turn replay or (B) replay with five-turn lookahead}
\begin{promptblock}
{
  "reasoning": null,
  "final_answer": [
    {
      "message_id": "item_1",
      "decision": "NO",
      "selected_turn": null
    }
  ]
}
\end{promptblock}

\paragraph{Answer template, (C) selection from the full transcript .}
\begin{promptblock}
{
  "reasoning": null,
  "final_answer": [
    {
      "message_id": "item_1",
      "decision": "PICK",
      "selected_turn": 1
    }
  ]
}
\end{promptblock}

\subsection{Candidate-message drafting}
\label{sec:app-prompt-draft}

Participants author messages with search and drafting support (Section~\ref{sec:study}). Optional blocks are included only when the corresponding input is present.

\paragraph{System message.}
\begin{promptblock}
You are an expert AI assistant embedded in a group conversation annotation tool. Your task is to generate a concrete, helpful agent utterance that could be inserted into the conversation to surface relevant external knowledge for the participants. The utterance should be natural, informative, and directly useful to the ongoing discussion. It should sound like a knowledgeable colleague contributing a fact, reference, clarification, or suggestion — not like a chatbot.
\end{promptblock}

\paragraph{User-message template.}
\begin{promptblock}
## Scenario
{scenario}

## Conversation transcript
{transcript_text}

{optional_user_selected_resources_or_search_results}

{optional_response_requirements}

{optional_current_draft_block}

{final_instruction}
\end{promptblock}

If participant-selected resources are available, the resources block is:
\begin{promptblock}
## User-selected resources (use these as primary sources)
1. **{title}**{optional_manual_source_note}
   {snippet up to 500 characters}
   URL: {url}
\end{promptblock}

If only search results are available, the resources block is:
\begin{promptblock}
## Relevant search results
1. **{title}**
   {snippet_or_abstract up to 500 characters}
   URL: {url}
\end{promptblock}

Optional contribution-type guidance:
\begin{promptblock}
## Response requirements
**Contribution type:** {type_guidance}
**Importance level:** {importance_label}
\end{promptblock}

Contribution-type guidance values:
\begin{promptblock}
heads_up: Introduce a useful fact, paper, example, or context about a topic the conversation has touched on but not yet considered.
pushback: Challenge or question something someone said in the conversation.
support: Confirm or back up something someone said in the conversation with evidence.
suggestion: Recommend what the group could do, consider, read, compare, or decide next.
\end{promptblock}

Importance labels:
\begin{promptblock}
importance >= 0.75: Essential or very important
importance >= 0.5: Important
importance >= 0.25: Somewhat important
otherwise: Nice to have
\end{promptblock}

Final instruction without an existing draft:
\begin{promptblock}
Based on the conversation and any search results above, generate a single concrete and helpful agent utterance that would benefit the group discussion. Keep the response under 100 words. Be concise and direct. Include inline citations for the sources provided above. For sources with URLs, use markdown hyperlinks: e.g. 'According to [Smith et al. (2023)](https://example.com/paper), ...'. For manual sources without URLs, cite them by title in bold: e.g. 'As noted in **Introduction to Algorithms**, ...'. Only cite sources that were provided above. Respond with only the utterance text, no extra commentary.
\end{promptblock}

Final instruction with an existing draft:
\begin{promptblock}
The user has already started writing the response shown above. Refine their draft into a concise agent utterance that preserves the original intent. Keep the response under 100 words. Be concise and direct. Include inline citations for the sources provided above. For sources with URLs, use markdown hyperlinks: e.g. 'According to [Smith et al. (2023)](https://example.com/paper), ...'. For manual sources without URLs, cite them by title in bold: e.g. 'As noted in **Introduction to Algorithms**, ...'. Only cite sources that were provided above. Respond with only the utterance text, no extra commentary.
\end{promptblock}

\subsection{Neutralization}
\label{sec:app-prompt-neutral}

Timing runs use the neutralized message (Section~\ref{sec:app-models}).

\paragraph{System message.}
\begin{promptblock}
You are a text neutralization assistant. Your task is to rewrite an AI agent utterance so that it preserves all factual and informational content but removes any conversational cues that tie it to a specific moment in a conversation.

Specifically, you must remove or rephrase:
- Discourse markers (e.g. 'Actually...', 'By the way...', 'So...')
- Turn-specific references (e.g. 'As you mentioned earlier...', 'Going back to what Alice said...')
- Response-to-question patterns (e.g. 'To answer your question...', 'That's a great point, but...')
- Conversation participant name references (e.g. 'As Bob noted...', '@Alice'). Only remove names of conversation participants, NOT names that appear in citations or references.
- Hedging or politeness tied to the conversation flow
- Turn numbers or placement references (e.g. 'responding to turn 5', 'after turn 3', 'at this point in the conversation'). Remove ALL references to specific turn numbers as they leak placement information.

INSTRUCTION-STYLE DETECTION: If the text is written as behavioral instructions for an AI (e.g. 'The AI should fact-check...', 'When the user says X, respond with...', 'Fact-check what was said') rather than a direct message, you MUST rewrite it as the concrete utterance the AI would actually send. Convert the instruction into the actual helpful message that would appear in the conversation.

IMPORTANT: You must preserve all academic citations (e.g. 'Smith et al., 2020', 'doi:10.1234/...'), bibliographic references, institution names (e.g. 'MIT', 'World Health Organization'), author names within citations, markdown hyperlinks (e.g. '[text](url)'), and any other sourcing information. These are factual content, not conversational cues.

The output should read as a standalone informative statement that could appear in any context. Preserve the exact factual claims, data, references, citations, and recommendations. Do not add or remove information.

Respond with ONLY the neutralized text, no commentary.
\end{promptblock}

\paragraph{User message.}
\begin{promptblock}
Please neutralize the following agent utterance:

---
{raw_text}
---
\end{promptblock}

\subsection{Firing-justification judge}
\label{sec:app-prompt-judge}

The replay reasoning trace analysis in Section~\ref{sec:results} uses the following judge.

\paragraph{System message.}
\begin{promptblock}
You label the reasoning behind a decision to add a message to a group conversation AT A PARTICULAR MOMENT.

The reasoner was shown a conversation up to one turn and asked a yes/no question: should this message be inserted here, now? It could not see what comes next. It said yes. You are given its justification.

The distinction that matters is between arguing that the CONTENT FITS and arguing that THIS MOMENT is right. They are different claims, and a justification can make one without the other.

LABELS:
- timing_argued: Argues that THIS MOMENT is the right one - that something just happened which makes now, rather than earlier or later, the time.
- relevance_only: Argues the message is useful, on-topic or a good fit here, without arguing that this moment is better than a later one.
- waiting_considered: Explicitly weighs holding off - names a later moment that might be better, or says the topic may develop further.
- urgency: Argues the window is closing: the conversation is about to move on, or a decision is about to harden.
- prerequisite_met: Argues a condition has now been satisfied that was not satisfied earlier - the topic has been raised, the question asked, the claim made.
- no_reason: States the decision with no justification.

Rules:
- "rationales" must contain ONLY exact label keys from the list above, verbatim and lowercase, e.g. "relevance_only". Never put a sentence, an explanation, or a quote in that array. Explanations belong in "evidence".
- Assign every label that applies. `relevance_only` and `timing_argued` are the important pair and are mutually exclusive.
- Assign `relevance_only` when the justification only establishes that the message is useful, on-topic, or answers something - without arguing that later would be worse.
- Assign `timing_argued` when it gives a reason THIS turn beats a later one.
- `no_reason` is exclusive.
- "evidence" is at most 20 words quoted from the justification.

Example of the required shape:
{"rationales": ["relevance_only"], "primary": "relevance_only", "argues_timing": false, "evidence": "directly answers the question about price"}

Return ONLY JSON:
{"rationales": ["..."], "primary": "...", "argues_timing": true, "evidence": "short quote"}
\end{promptblock}

\paragraph{User message.}
\begin{promptblock}
CONVERSATION, UP TO AND INCLUDING TURN {checkpoint}:
{conversation_excerpt}

MESSAGE TO ADD:
{message_text}

DECISION: insert it here, after turn {checkpoint}

JUSTIFICATION:
{justification}
\end{promptblock}

\end{document}